\documentclass{article}

\usepackage[final,nonatbib]{nips_2016}

\usepackage[utf8]{inputenc} 
\usepackage{color}
\usepackage[T1]{fontenc}    
\usepackage{hyperref}       
\usepackage{url}            
\usepackage{booktabs}       
\usepackage{amsfonts}       
\usepackage{nicefrac}       
\usepackage{microtype}      
\usepackage{amsmath}
\usepackage{amssymb}
\usepackage{tikz}
\usepackage{subcaption}
\usetikzlibrary{fit,positioning}

\vspace{-5cm}
\title{Transfer Learning via Latent Factor Modeling\\ to Improve Prediction of Surgical Complications}
\vspace{-1cm}
\author{
  Elizabeth C Lorenzi\\
  Department of Statistical Science\\
  Duke University\\
  Durham, NC 27705 \\
  \texttt{elizabeth.lorenzi@duke.edu} \\
   \And
  Zhifei Sun \\
  Department of Surgery\\
  Duke University\\
  Durham, NC 27705 \\
  \texttt{zhifei.sun@duke.edu}\\
   \And
  Erich Huang \\
  Department of Bioinformatics\\
  Duke University\\
  Durham, NC 27705 \\
  \texttt{erich.huang@duke.edu}\\
 \And
  Ricardo Henao\\
  Department of Electrical and Computer Engineering\\
  Duke University\\
  Durham, NC 27705\\
  \texttt{ricardo.henao@duke.edu}
  \And
  Katherine A Heller \\
  Department of Statistical Science\\
  Duke University\\
  Durham, NC 27705 \\
  \texttt{katherine.heller@duke.edu}\\
}

\begin{document}
\maketitle

\vspace{-1cm}
\section{Introduction}
Currently, surgical complications arise in $15\%$ of all surgical procedures performed, and can increase to $50\%$ or more in high risk surgeries \cite{healey2002complications}. Surgical complications are associated with decreased quality of life to patients, increased risk of litigation to surgeons, and significant costs to the health system.  Data from University of Michigan have shown that the cost of a major complication is over $\$11,000$ dollars per event \cite{dimick2004hospital}. Additionally, because complications frequently require further hospitalization, the total cost of care can multiply by a factor of five as the length of stay increases. 
 
To this end, the American College of Surgeons (ACS) created the National Surgical Quality Improvement Program (NSQIP), a national registry of patient preoperative clinical variables and postoperative outcomes. Over 700 hospitals with varied geographic, socioeconomic, and patient case-mix backgrounds participate in NSQIP, providing blinded information to all NSQIP hospitals. However, while NSQIP data provides a rich source of information, key pieces of information are missing in these data, including hospital-specific covariates describing how clinical care differs among health systems. While NSQIP contains 971,455 cases collected nationally from 2005-2008, our own university hospital only contributed  13,493 cases, and these cases have significantly different rates of outcomes than the national average. Predictive models of all the national NSQIP data represent an aggregated effect from all hospitals, limiting accuracy and utility for individual hospitals.

In an effort to leverage a larger source of data while creating accurate predictions for individual hospitals, we consider a transfer learning method that adjusts for the discrepancies between the multi-hospital national population and a hospital-specific population. Transfer learning aims to transfer knowledge between task domains to enrich the testing environment of interest \cite{pan2008transfer}. By learning and adjusting for the distributional differences of source (national) and target (local) data, we can use the additional information in the source data to better inform predictions in the target domain. 

We develop a hierarchical latent factor model to perform this task. This hierarchical model learns a set of latent factors that explain the dependence structure among the predictors while taking into account differences between populations.

\section{Transfer Learning Latent Factor Model (TL-LFM)}
Consider the set of labeled predictors, $\{X^{T}_i: i = 1, ...,n_T\}$ and $\{X^{S}_i: i = 1, ...,n_S\}$, that represent the pre-operative and intra-operative covariates for patients in the target and source data, respectively. The target data refers to the local hospital or the target distribution for predictions. The source data will refer to the multi-hospital NSQIP data. Each patient has a corresponding outcome vector, representing the resulting outcomes of their surgical procedure, represented as  $\{Y^{T}_i: i = 1, ...,n_T\}$ and $\{Y^{S}_i: i = 1, ...,n_S\}$, for the target and source data respectively.  We also use the combined notation for simplicity, $\{X_i: i = 1, ...,n\}$, where $n=n_T + n_S$.  $X$ contains $P$ variables, where $j = 1,...,P$. 

In traditional learning, we train a model with $\{X^T, Y^T\}$ to predict new patients within the target hospital. With the availability of additional related data, we may pool together the samples and train on $\{X, Y\}$ to predict for the target hospital. Though pooling is beneficial in many cases, we ignore the differences of the source data to the target population. Our goal is to utilize all data available, while taking into account these distributional differences.

We relate the combined set, $X$, to an underlying $k$-vector of random variables, $f_i$, using a standard $k$-factor model \cite{lopeswest}.
\begin{equation}
X_i = \beta^{t_i} f_i + \epsilon_i
\label{Eqn1}
\end{equation}
where $t_i$ $\in \{S, T\}$, representing the membership of the patient in the source or target data. $\beta^{t_i}$ represents the $P\times K$ factor loadings matrix, learned separately for the two populations, the factors, $f_i$ are independent and distributed with $f_i \sim N(0, I_K)$, and $\epsilon_i$ are independent and distributed $\epsilon_i \sim N(0, \Sigma)$. $\Sigma$ is a diagonal matrix that reduces to a set of $P$ independent inverse gamma distributions, with $\sigma_j^2 \sim \text{IG}(v/2, v/2)$ for $j= 1,...,P$. 

Due to large imbalances between the size of the source and target data, we set the following priors on the factor loadings to facilitate sharing of information.
\[m_j \sim \text{N}(0, \frac{1}{\phi}I_k), \quad
\beta^{S}_j \sim N(m_j, \frac{1}{\phi_{S}} I_K), \quad \beta^{T}_j \sim N(m_j, \frac{1}{\phi_{T}} I_K)\]
\vspace{-.4cm}

We learn a global level loadings matrix, $m$, and center the prior for the $j$th row of $\beta^T_j$ and $\beta^S_j$ on the draw from $m_j$. This results in a more informed prior for the target data, while still allowing flexibility for it to adapt to the population. Depending on the severity of the class imbalance between the source and target and the level of differentiation between the two groups, it may improve the performance of the model to tighten or expand the precision on the $\beta^T_j$ draw. Alternatively, we can set a prior on $\phi^{t_i}$ and learn the appropriate adjustment. 

Conditioned on the factors, the observed variables are uncorrelated. Dependence between these observed variables is induced by marginalizing over the distribution of the factors, resulting in the marginal distribution, $X_i \sim N_p(0, \Omega^{t_i})$ where $\Omega^{t_i} = V(X_i|\beta^{t_i}, \Sigma) = \beta^{t_i} \beta^{t_i \prime} + \Sigma$. Consider an equivalent prior specification of the factor loadings matrix,  $\beta^{t_i}_j =  m_j + \epsilon^{t_i}$, where $\epsilon^{t_i} \sim N(0, \frac{1}{\phi^{t_i}} I_K)$. It follows that the marginal distributions for the two groups are, $X^{t_i} \sim N(0, (m+\epsilon^{t_i}) (m+\epsilon^{t_i})^\prime + \Sigma)$. Through this specification, we see that the covariance structure for each population is a composition of a shared covariance plus a population specific adjustment. This facilitates sharing between the populations, while still allowing for an adjustment to more accurately portray the target's covariance. It also aids in learning the target groups' covariance structure despite the much smaller size of their data. The model results in the common factors explaining all of the dependence structure among the $P$ covariates for $X$.  

The predictors, $X$, are mixed data, containing both binary and continuous data. Therefore, we extend the factor model above to binary data. In the binary model, we use a probit link to learn $P(X_{ij}=1|f_i, \beta^{t_i}_{j}) = \Phi(\beta^{t_i}_{j} f_i)$, where $j$ is the $j$th binary column of the predictors matrix, and $\Phi$ represents the CDF of a normal distribution \cite{albert1993bayesian}. The probit link offers a convenient inference implementation using data augmentation. 

\subsection{Latent Factor Regression}
Next we relate the learned factors to the surgical outcomes. We will measure the outcome of any morbidity or complication after surgery, but our framework easily extends to multivariate outcomes. For each $X_i$, we have a corresponding response, $Y_i$, coded either 0 or 1. Let $Z$ = \{Y, X \} represent the full data, where $Z^{T}$ and $Z^{S}$ represents the data for each subgroup. We then jointly model the $Z$s using (\ref{Eqn1}). 

The posterior predictive distribution then is easily found by solving, $f(y_{n+1}|z_1,...,z_n, x_{n+1}) = \int f(y_{n+1}|x_{n+1}, \Theta) \pi(\Theta|z_1,...,z_n) d\Theta$. The joint model implies that $E(y_i|x_i) = x_i^\prime \theta_{t_i}$ with covariance matrix $\Omega_{t_i YX}$. The resulting coefficients, $\theta_{t_i} = \Omega^{t_i -1}_{XX} \Omega^{t_i}_{YX}$, are found by correctly partitioning the covariance matrix, $\Omega_{t_i}$. This then results in the true regression coefficients of $Y$ on $X$ for both target and source data.

\subsection{Inference}
The latent factor model proposed is fully conjugate, allowing for easy implementation of Markov Chain Monte Carlo (MCMC) via Gibbs sampling for posterior estimation of the latent variables,$\{F, Z^*\}$, and model parameters $\Theta = \{\beta^T, \beta^S, \Sigma \}$. In \cite{gewekezhou96}, \cite{polasek1997factor}, \cite{aguilar2000bayesian}, and \cite{lopeswest}, they provide details of the MCMC algorithm for the k-factor model.

\section{Experiments}
The development of this transfer learning model is motivated by our goal of accurately predicting the risk of adverse outcomes for individual patients at our local hospital.  We work with data from ACS NSQIP, a national surgery program collected from participating hospitals across the country. The data contain information about patients' demographic, medical history, lab results, procedural information via CPT codes, and postoperative outcomes.

\subsection{Simulation Studies}
To demonstrate the effectiveness of our method, we simulate scenarios in which we have two populations of differing size with differing amounts of variation between them.  We test our method for different ratios of source to target data and repeat each experiment 10 times. For each experiment, we begin with a sample of 5000 patients, of which 1000 come from the target distribution. The first experiment trains on 2800 of the national patients and 700 from the single-hospital population, and the second and third experiment use the ratios of multi-hospital to single hospital of 2500:500 and 4000:200. We compare the results of our transfer learning latent factor model (TL-LFM) to a factor model (LFM) with no hierarchy (as in \cite{lopeswest}) and to the lasso penalized regression \cite{lasso}.

We simulate $Z_i$, for $i=1,...,5000$ from a 70-dimensional normal distribution, with zero mean and covariance equal to $\Omega^{t_i} = \beta_{t_i}\beta_{t_i '} + \Sigma$. We choose $K$ to be 20 and sample the rows of the global loadings matrix, $m_j$, from Normal$(0,I_k)$. We then draw each row of $\beta^{S}_j$ and $\beta^T_j$ from a Normal$(m_j, I_k)$. We draw the diagonal of $\Sigma$ from an InvGamma(1, 0.5) with prior mean equal to 2.

\begin{figure}[h]
\vspace{-.5cm}
\centering
\begin{subfigure}{0.3\textwidth}
\includegraphics[width=0.8\linewidth]{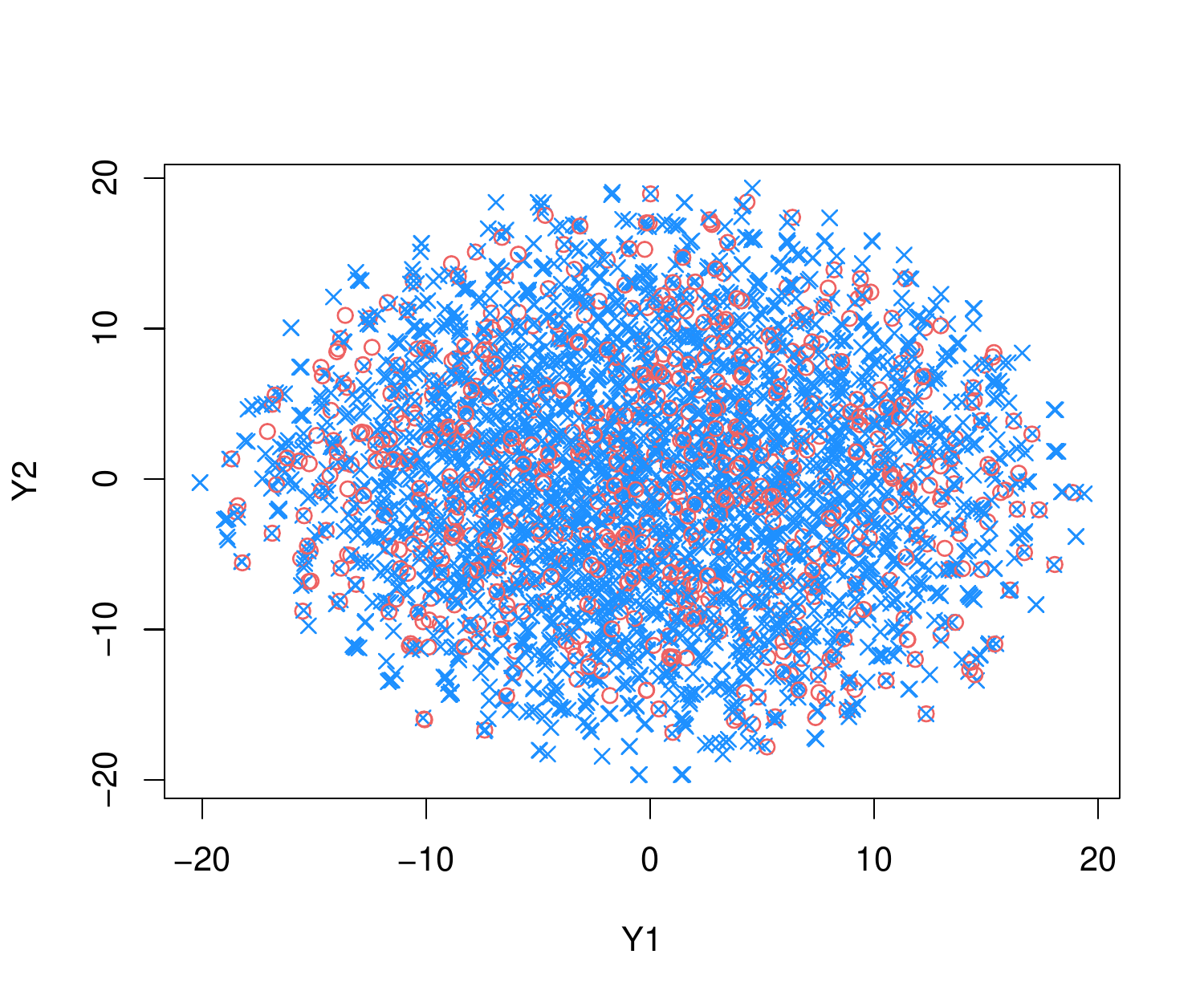} 
\caption{TL - 700:2800}
\label{fig:subim1}
\end{subfigure}
\begin{subfigure}{0.3\textwidth}
\includegraphics[width=0.8\linewidth]{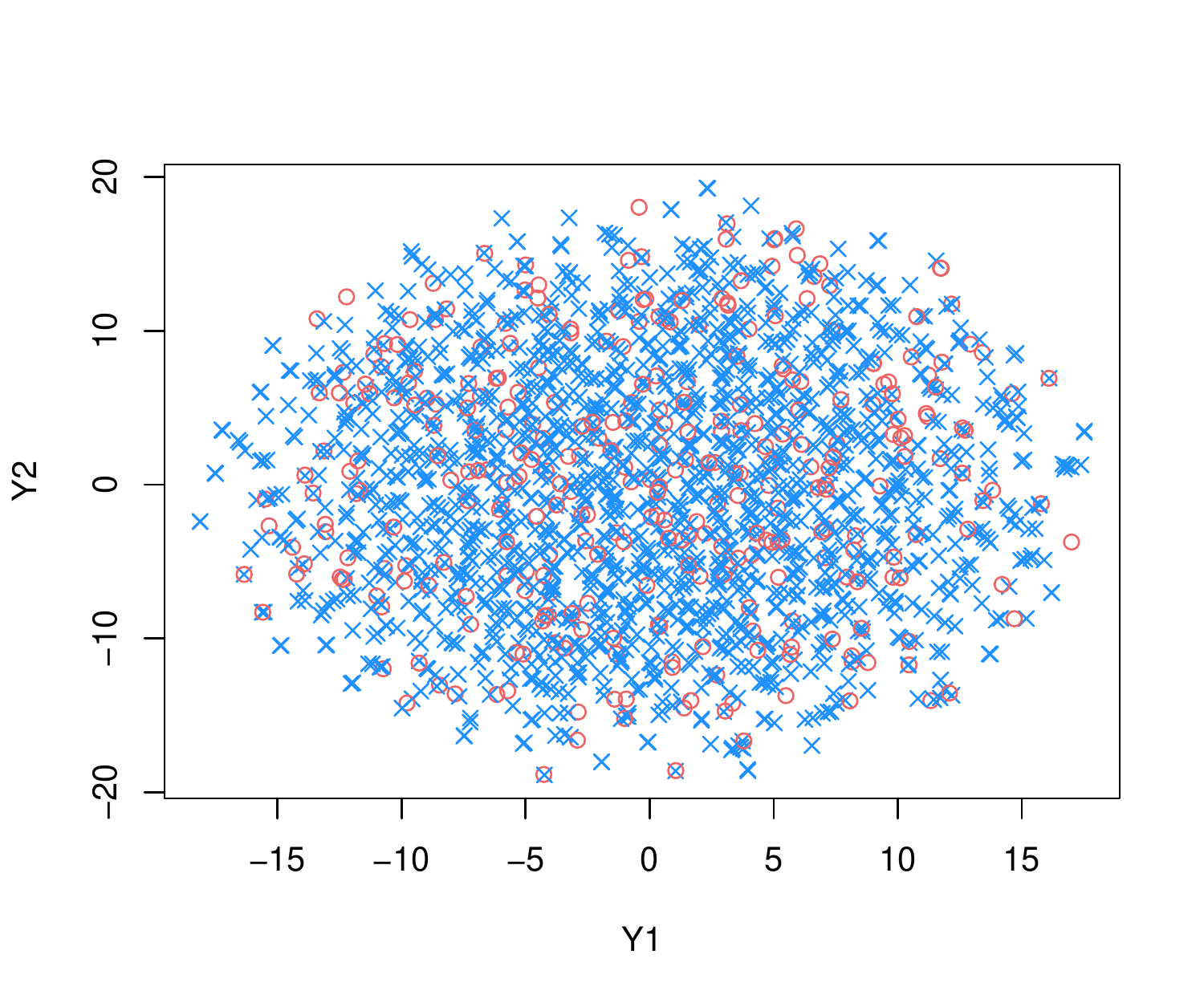}
\caption{TL - 500:2500}
\label{fig:subim2}
\end{subfigure}
\begin{subfigure}{0.3\textwidth}
\includegraphics[width=0.8\linewidth]{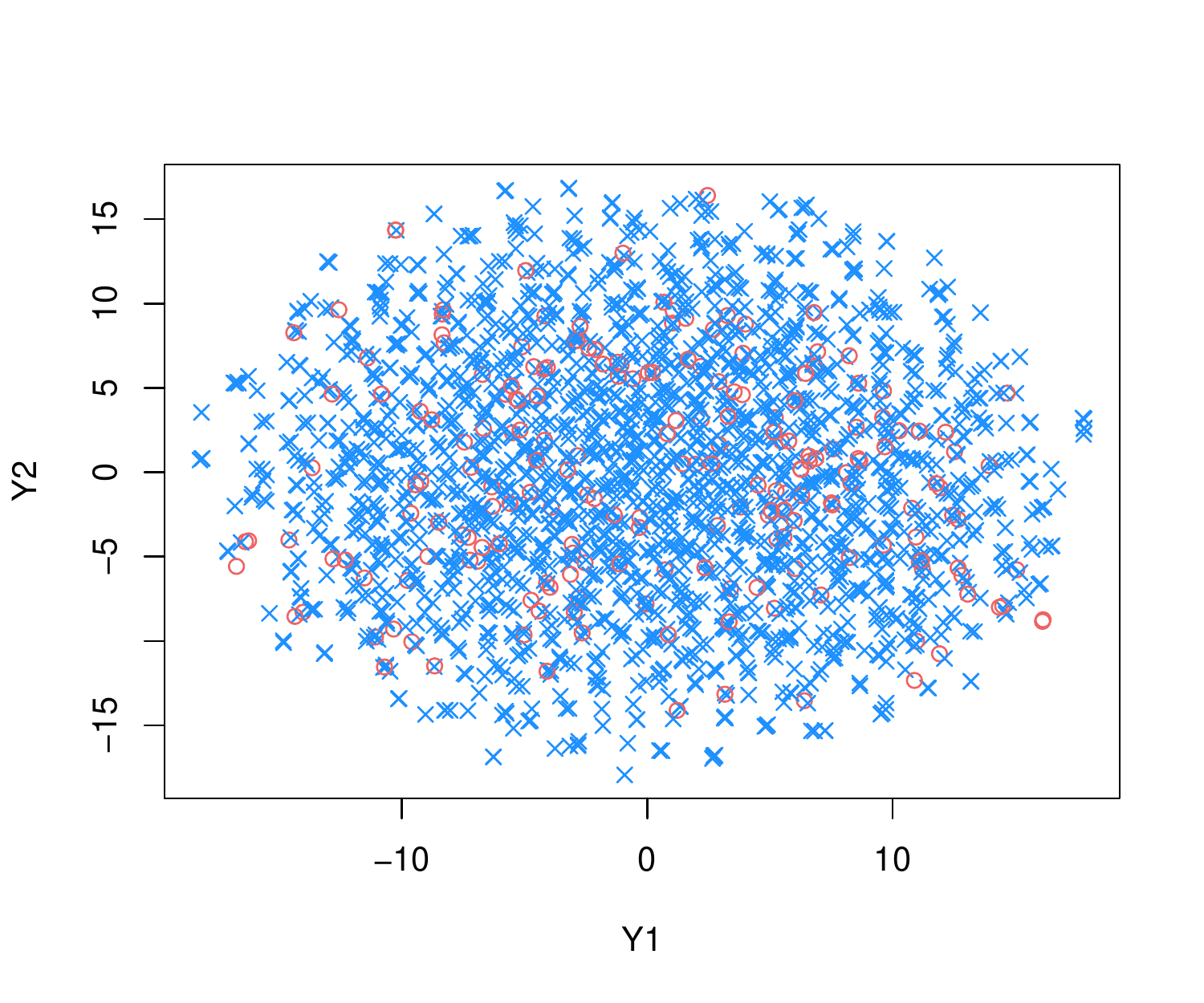}
\caption{TL - 100:2000}
\label{fig:subim3}
\end{subfigure}
 \begin{subfigure}{0.3\textwidth}
\includegraphics[width=0.8\linewidth]{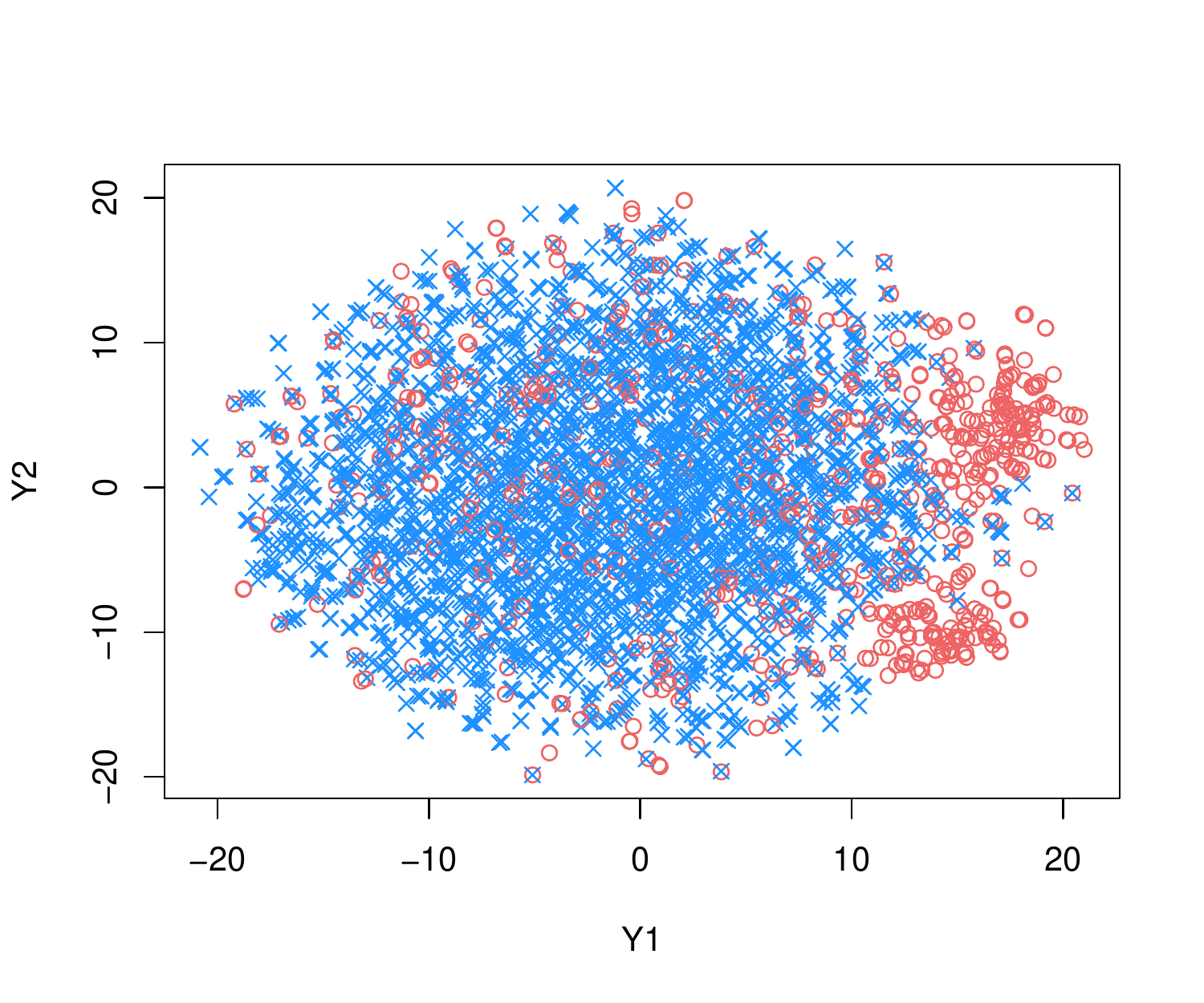} 
\caption{NoTL - 700:2800}
\label{fig:subim1}
\end{subfigure}
\begin{subfigure}{0.3\textwidth}
\includegraphics[width=0.8\linewidth]{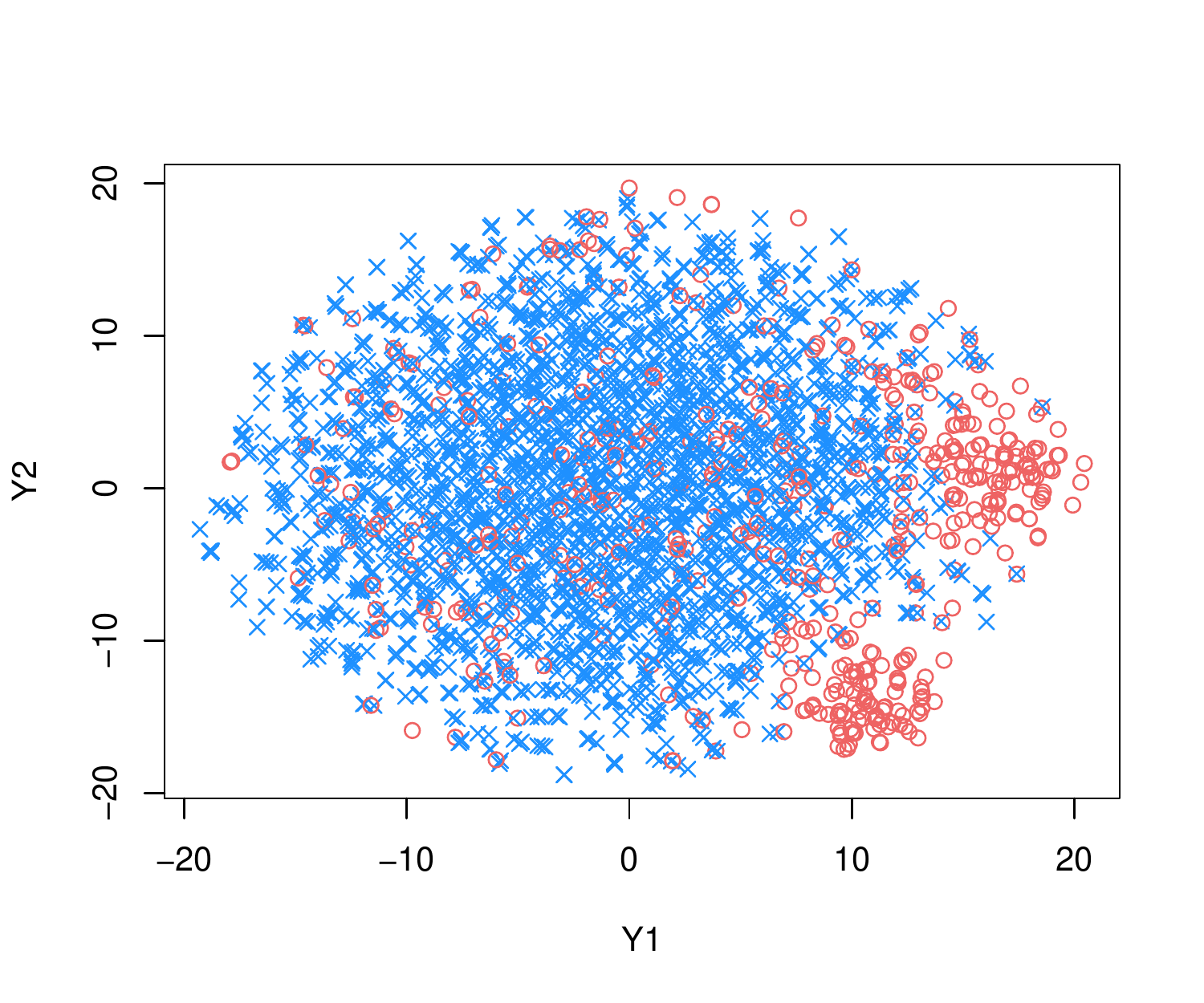}
\caption{NoTL - 500:2500}
\label{fig:subim2}
\end{subfigure}
\begin{subfigure}{0.3\textwidth}
\includegraphics[width=0.8\linewidth]{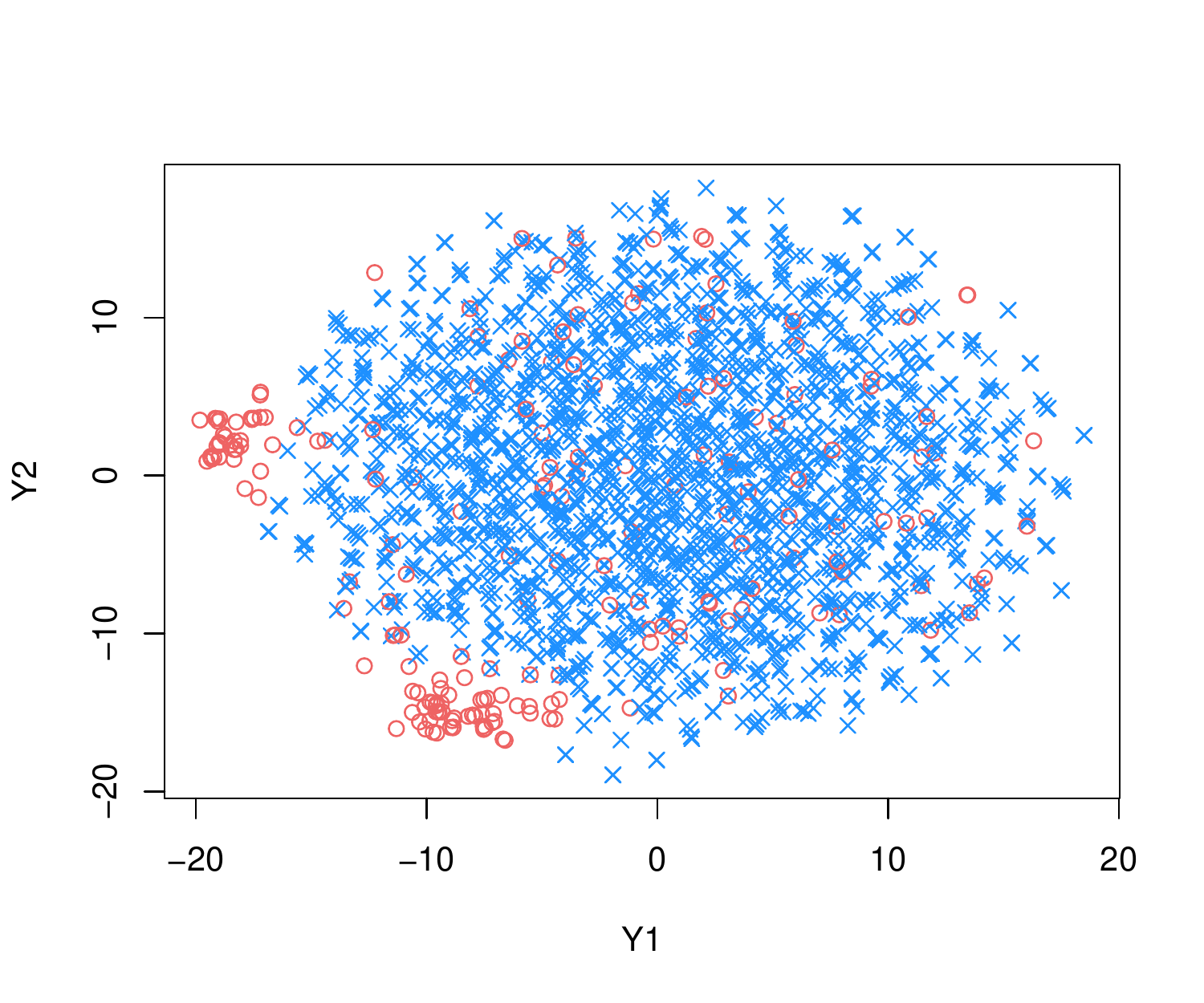}
\caption{NoTL - 100:2000}
\label{fig:subim3}
\end{subfigure}
\caption{Plots of tsne-representation of the learned latent factors from simulation study using different ratios of target to source patients (e.g. 700:2800 represent 700 in the target and 2800 in the national). TL : factors from the latent factor model with transfer learning and NoTL: latent factor model with no transfer learning. (Blue $\times$- Source, Red $\circ$-Target)}
\label{factors}
\end{figure}

In Figure \ref{factors}, we display the last iteration of the factor scores learned in our model, colored and marked on the plot to identify the source (blue crosses) from the target data (red circles). We project the 20-dimensional factor scores to a 2-dimensional space using t-SNE \cite{tsne}, and use this projection to compare the TL-LFM from LFM. We aim to appropriately model the differences in our populations through the loadings matrix such that the factors themselves represent the underlying dependence structure of all patients, regardless of population. Therefore the factors serve as a new representation of the data that make the populations "similar" and is free of their distributional differences. The top set of plots (a-c) displays two groups that are indistinguishable from each other in the feature space. The bottom set of plots (d-f) show two separate distributions, where the target data seem to occupy certain regions of the source data's feature space. When we do not adjust for the data sources, the latent variables try to learn the differences between them and therefore do not achieve our goal of finding a representation that eliminates the distributional variations. By resulting in a representation that adjusts for these differences, it alerts us that we are properly modeling the populations.

Lastly, we test the performance of these models using a held out test set and learning the posterior predictive distribution. We calculate the area under the ROC curve (AUROC) and the standard error of the AUROC from the 10 samples in each experiment. We show the results of the models for both a held-out target test set as well as a held-out source test set. Our TL-LFM performs best when more target data is available, as in the 700:2800 and 500:2500 ratio experiments. However, when we greatly increase the discrepancy as in the 200:4000 experiment, we no longer outperform Lasso.

\begin{table}[ht]
\centering
\begin{tabular}{|r|r|r|r|r|r|r|}
  \hline
  & TL-LFM (T) & TL-LFM (S) & LFM (T) & LFM (S) & Lasso (T) & Lasso (S) \\ \hline
 700:2800 & 0.81 (0.005) & 0.80 (0.006) & 0.60 (0.005) & 0.76 (0.006) & 0.73 (0.004) & 0.78 (0.005) \\ 
 500:2500 & 0.80 (0.009)  & 0.79 (0.011) & 0.55 (0.007) & 0.76 (0.007) & 0.60 (0.005) & 0.69 (0.005) \\
 200:4000 & 0.64 (0.025)  & 0.66 (0.025) & 0.50 (0.017) & 0.67 (0.025) & 0.65 (0.024) & 0.66 (0.025) \\
   \hline
\end{tabular}
\vspace{1mm}
\caption{Results from 3 simulation experiments. We report the AUROC with standard errors calculated from ten samples. We show prediction results for different held-out test sets: T- a target only test set, S - a source only test set.}
\vspace{-8mm}
\end{table}

\subsection{Surgical Complications results}
\textbf{Data}: Our dataset is comprised of 984948 patients, 971455 of which are from the NSQIP dataset, and 13493 are contributed from our local health system collected between 2005 and 2009. For each patient, we have recorded a collection of preoperative variables and postoperative outcomes. There are 75 predictor variables, including continuous variables, such as lab values, age, BMI, as well as binary variables, indicating varying information about the patient history. Many of the binary variables are highly imbalanced, the majority of them with means less than $5\%$. We aim to predict whether the patient will have any complication within 30 days of their surgery.

We train our models using large subsets of the data, N=110,000, where 10,000 of those samples are from our local health system. We then test on the remainder of the target data. We select the number of factors, $K$, by using 5 fold cross validation, evaluating the prediction performance with AUROC. Our TL-LFM outperforms a LFM but does not outperform Lasso. The AUROCs are 0.73, 0.60, and 0.76 for TL-LFM, LFM and Lasso, respectively.

\section{Future Directions}
The simulation study provides evidence that the latent factor framework presented achieves our two goals of using all data available and improving prediction on target data. However, the simple structure of our model proves inflexible when we move to more complicated data problems, as in our surgical complications data. The problems plaguing our model fit on the surgical data are due to two reasons. First, the target and source populations overlap greatly and the differences between these data are found in specific areas of the feature space, and second, the data are highly correlated and are sparse. We therefore look to improve this model by developing a more sophisticated factor model that models modes of the distribution and introduces sparsity into the model. There are papers in the literature for sparsity inducing factor models, such as \cite{sparseinfinitefactor}, \cite{knowles2007infinite},\cite{west2003}, or through sparse priors as reviewed in \cite{polson2010shrink}. Due to the multi-modality of the data that we can observe with $t$-sne, we also look to allow for further decomposition of the covariance structure using ideas from mixture models. 



Specifically, we are developing an infinite factor model that combines the "rich gets richer" properties from the Hierarchical Dirichlet Process \cite{Teh2012} with our original latent factor framework. This framework, most often used for hierarchies of mixture models, is used in this case to provide a hierarchical weighting scheme to the factor model via the stick-breaking process. We allow the number of factors to approach infinity, and adjust their weights through the loadings matrix using the mixing proportions learned from the DP. This induces sparsity into our model, allows each population to make up different compositions of the factors, and results in strong sharing across populations.

\section{Discussion}
Our transfer learning latent factor model provides a novel framework for incorporating transfer learning in a latent factor regression model by learning and adjusting for the population differences between source and target data. We faced limitations in the initial model with beating the performance of lasso on the surgical data. We attribute this to the strong correlation and sparsity inherent in our data as well as the restrictions we assume within the current factor model framework of how information is shared between populations. A model allowing each population to be made up of different compositions of factors is better suited for populations that differ in small idiosyncratic ways. These are our focus for future work as outlined in 4. We envision developing the methods presented here further as different types of EHRs become available and new structure in this data requires more advanced modeling. 

Our work is a part of the continued effort to create a clinical platform to deliver individualized risk scores of complications at our university's health system. We plan to implement this framework directly into their electronic health system, so that clinicians will be able to assess the predicted complications directly through the patient's chart and treat the patient with suggested interventions that address the patient's increased risk.

\small

\bibliography{nips_2016}{}
\bibliographystyle{plain}
\end{document}